\documentclass{article} % For LaTeX2e
\usepackage{iclr2025_conference,times}
\usepackage{enumitem}  % 支持自定义列表样式
\usepackage{amsmath}   % 支持数学公式环境
\usepackage{amsthm}
\theoremstyle{plain}
\newtheorem{theorem}{Theorem}[section]

\theoremstyle{definition}
\newtheorem{definition}[theorem]{Definition}

\theoremstyle{remark}

\usepackage{hyperref} 
\usepackage{url}
\usepackage{graphicx}
\usepackage{subcaption}
\usepackage{graphicx}  % 引入图片宏包
\usepackage{booktabs}

\usepackage{amsmath,amsfonts,bm}

\def\eqref#1{equation~\ref{#1}}
\def\1{\bm{1}}

\DeclareMathAlphabet{\mathsfit}{\encodingdefault}{\sfdefault}{m}{sl}
\SetMathAlphabet{\mathsfit}{bold}{\encodingdefault}{\sfdefault}{bx}{n}

\title{Pinpointing  crucial steps: Attribution-based Credit Assignment for Verifiable Reinforcement Learning
}
\author{Haisen Luo \thanks{Equal contributions,order is determined by flip of coin.} \& Zhenyu Li\footnotemark[1] \& Yihua Liu \footnotemark[1], Junxi Yin \thanks{Corresponding Author.} ,  Dan Liu ,Zequn Li , Xiaohang Xu\\
Institute of Artificial Intelligence\\
Taikang Insurance Group Inc\\
\texttt{westlong4ai@gmail.com},\\
\texttt{leestar127@gmail.com},\\
\texttt{liuyihua1994@gmail.com}, \\
\texttt{curtisyin@gmail.com},\\
\texttt{liudan920521@gmail.com}, \\
\texttt{\{lizq113,xuxh68\}@taikanglife.cn}
}

\iclrfinalcopy % Uncomment for camera-ready version, but NOT for submission.
\begin{document}

\maketitle
\begin{abstract}
While Reinforcement Learning with Verifiable Rewards (RLVR) enhances complex reasoning in LLMs, current methods struggle to balance exploration and exploitation. This leads to critical issues like inaccurate credit assignment for intermediate steps and premature entropy collapse, limiting model performance. To address this, we introduce Attribution-based Contribution to Policy Optimization (ACPO), a phased framework that incorporates a difficulty-aware curriculum. ACPO improves exploration by using trajectory semantic segmentation and an attribution-based representation to dynamically regulate policy entropy, thus mitigating its collapse. Concurrently, it enhances exploitation with a factorized reward system that precisely quantifies the hierarchical contribution of each reasoning step, ensuring accurate credit assignment. Extensive experiments on challenging benchmarks, including AIME, MATH, and AMC, demonstrate that ACPO significantly outperforms existing state-of-the-art approaches.
\end{abstract}

\section{Introduction}

In recent years, there has been an explosive growth in the demand for applying Large Language Models (LLMs) to complex cognitive tasks such as mathematical reasoning, logical proof, and multi-step decision-making. This has driven the evolution of model training paradigms—shifting from Supervised Fine-Tuning (SFT), which relies on annotated data, toward more generalizable Reinforcement Learning (RL) frameworks. A particularly promising direction within this domain is Reinforcement Learning with Verifiable Rewards (RLVR), which has gained widespread adoption over methods like Reinforcement Learning from Human Feedback (RLHF) \cite{li2025miromind, deepseek2025deepseekr1}. The advantages of RLVR lie in its lower annotation costs and its ability to minimize the introduction of subjective human biases by sourcing rewards from objective, automated verifiers, such as code compilers or mathematical theorem provers \cite{xin2025deepseekprover}.

% A representative and foundational RLVR algorithm is GRPO \cite{shao2024deepseekmath}, which provides a simple yet effective approach based on "outcome supervision"—that is, rewards are determined solely by the correctness of the final answer. While powerful, this paradigm reveals inherent limitations when applied to tasks requiring long and intricate reasoning chains. The sparse reward signal fails to provide fine-grained guidance for the model's intermediate steps, creating a critical credit assignment problem: when a trajectory succeeds or fails, it is difficult to determine which specific steps were responsible. This often leads to inefficient learning, where entire "poor-performing" sequences are penalized without targeted correction. Furthermore, this approach, coupled with other factors like premature entropy collapse, creates a ceiling on the model's ultimate reasoning capabilities.

Recent RLVR  research can be broadly categorized along three primary axes: policy optimization, verification paradigms, and exploration mechanisms.

\textbf{Policy Optimization.} Following the groundwork laid by GRPO \cite{shao2024deepseekmath}, numerous algorithms have been proposed to enhance training stability and performance. Dr.GRPO provided theoretical guarantees by deepening the understanding of ordinal reward design \cite{liu2025drgrpo}. DAPO improved the exploration upper bound by rectifying the PPO clip strategy and refining token-level reward allocation \cite{yu2025dapo}. For sequential tasks, GSPO introduced a grouped policy update rule tailored for multi-step verification \cite{zheng2025gspo}, while GMPO used a geometric mean function to balance process and outcome verification rewards \cite{zhao2025gmpo}. DCPO further optimized update stability in large-scale scenarios with an adaptive strategy \cite{yang2025dcpo}, and REINFORCE++ struck a balance between PPO and GRPO by focusing on token-level rewards to reduce the coupling between verification and policy updates \cite{hu2025reinforcepp}.

\textbf{Verification Paradigms.} To overcome the reward sparsity of outcome-only supervision, a major thrust of research has been the development of "process-based rewards." This paradigm, exemplified by works such as rStar-Math \cite{guan2025rstarmath}, PRIME \cite{cui2025prime}, and Beyond the Last Answer \cite{hammoud2025beyond}, often employs Process Reasoning Models (PRMs) or step-scorers to annotate the logical validity of intermediate reasoning steps. While effective at providing denser feedback, these methods require significant additional training resources and can be susceptible to reward hacking. Other research, including DeepSeek-Prover-V1.5 \cite{xin2025deepseekprover} and MiroMind-M1 \cite{li2025miromind}, has expanded the diversity of automated verifiers, providing a rich toolkit for reward design.

\textbf{Exploration Mechanisms.} A critical challenge in RLVR is preventing "entropy collapse," where the policy model prematurely converges to a narrow set of reasoning strategies. Recognizing this, researchers have investigated methods to explicitly encourage exploration. Works such as The Entropy Mechanism of Reinforcement Learning \cite{cui2025entropy} and First Return, Entropy-Eliciting Explore \cite{zheng2025fr3e, eco_et_al_2018} have demonstrated that incentivizing policy entropy prompts models to conduct broader searches, leading to the discovery of higher-quality solutions. Further research has revealed that high-entropy minority tokens can be particularly effective in guiding the policy to capture fine-grained verification information, providing a theoretical basis for advanced reward design in RLVR \cite{wang2025highentropy}. Collectively, these research threads, supported by system-level implementations like DeepSeek R1 \cite{deepseek2025deepseekr1} and DAPO \cite{yu2025dapo}, form the technical foundation upon which our work builds.

While powerful, these paradigm reveals inherent limitations when applied to tasks requiring long and intricate reasoning chains. The sparse reward signal fails to provide fine-grained guidance for the model's intermediate steps, creating a critical credit assignment problem: when a trajectory succeeds or fails, it is difficult to determine which specific steps were responsible. This often leads to inefficient learning, where entire "poor-performing" sequences are penalized without targeted correction. Furthermore, this approach, coupled with other factors like premature entropy collapse, creates a ceiling on the model's ultimate reasoning capabilities.

These challenges highlight two fundamental, intertwined problems in the field:\textbf{ 1)  Achieve precise, step-level credit assignment without incurring the high costs and potential of manually annotated process-based rewards. 2) Systematically manage the trade-off between exploration and exploitation to prevent the model from converging to suboptimal reasoning paths.} Existing methods have made incremental progress on these fronts, but a unified framework that holistically addresses both issues remains elusive. There is a clear need for a method that can "surgically" attribute credit to critical reasoning steps while dynamically guiding the model's search for novel solutions.

To address these fundamental challenges, we introduce Attribution-based Contribution in Policy Optimization (ACPO), a novel two-stage algorithmic framework. ACPO is designed to fundamentally improve both exploitation and exploration in RLVR. For exploitation, it introduces a factorized reward system that ensures accurate credit assignment. By using trajectory semantic segmentation and an attribution-based representation, ACPO can precisely quantify the hierarchical contribution of each reasoning step to the final outcome. For exploration, it employs a difficulty-aware curriculum and a multi-stage, covariance-based decision-making process to dynamically regulate policy entropy. This mitigates premature entropy collapse and guides the model to discover more diverse and effective reasoning paths. Our contributions are a framework that resolves the credit assignment problem with fine-grained, step-level attribution and a principled approach to optimizing the exploration-exploitation balance in complex reasoning tasks.

\section{Methods}

% \subsection{Data Constructions(put it in experiment section)}

% In terms of data, we integrated guided data and raw data. The specific process is as follows: First, we leverage the LLM that we would like to train to perform reasoning on the dataset, and define the reasoning difficulty based on the number of successful rollouts. Subsequently, for data with higher difficulty, we conduct multiple rollouts, select reasoning trajectories that yield correct outcomes, and extract questions and prefixes from these trajectories to construct new sample pairs. This approach enables dataset fusion guided by an expert knowledge base.

\subsection{Step-wise Attribution Credit Assignment}

% 插入图片
\begin{figure}[htbp]
    \centering
    % 宽度设为文本宽度的60%，保持比例
    \includegraphics[width=1.0\textwidth, keepaspectratio]{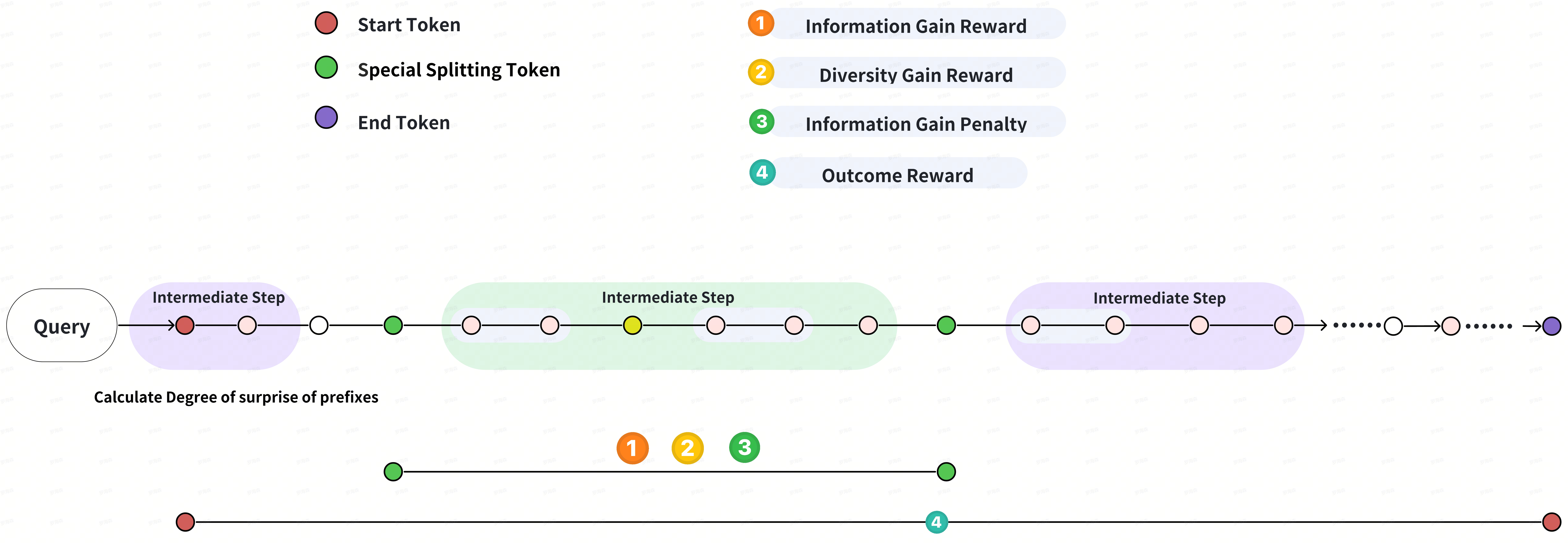}
    \caption{Spliting strategy}  % 标题
    \label{fig:sample}  % 标签
\end{figure}

To address these shortcomings, we have designed a step-wise causal reward assignment mechanism. This mechanism focuses the model’s optimization on the causal relationship between generated outputs and reasoning steps. By leveraging step-wise optimization objectives and precise credit assignment, we guide the model to learn asymptotically. This enables our model to filter out redundant information, efficiently converge toward objectives, and simultaneously break through strategic bottlenecks. The specific method is as follows:

\subsubsection{Dynamic Step Segmentation Strategy}

Traditional rule-based methods for segmenting reasoning trajectories are often rigid, coarse, and struggle to adapt to complex tasks.To address these limitations, we propose an adaptive segmentation strategy that leverages the model's token-generation probabilities. Our approach identifies key decision points by focusing on high-entropy tokens, which often correspond to crucial junctures in the reasoning process.

% High-entropy tokens serve to guide reasoning (e.g., "first", "because") and support self-verification/correction. A small number of "branching tokens" (high-entropy tokens) act as key decision points in the reasoning path. We consider segmenting trajectories directly using high-entropy nodes. Specifically, the probability of generating a token is calculated as:

\begin{equation}
c_{s_i^n}=p\left(s_i^n \mid \pi, q, s_{<i}^n\right)
\end{equation}

We segment different steps using segmentation markers, and further refine the segmentation via high-entropy tokens and supplementary rules. The specific process is as follows.

% \textbf{Extract the marking words based on entropy}First, we extract the top 5\% tokens with high entropy or significant probability changes from the confidence scores and samples, treating them as segmentation candidate nodes. From these candidates, we select nodes that mark the boundaries of reasoning steps—specifically tokens with explicit transitional meanings (e.g., "however", "thus", "so", "but", "wait"). These tokens indicate the completion of a reasoning phase by the model (e.g., problem analysis, formula derivation, conclusion verification).
% Segmentation with constraints: After selecting the candidate tokens, we perform segmentation while satisfying two constraints: "maintaining a minimum interval between segments" and "aligning with complete sentences". This avoids over-dense segmentation—meaning no segmentation occurs within a sentence, and multi-ach resulting step is a coherent and meaningful unit of reasoning.

Our segmentation process is as follows. First, we identify candidate segmentation points by \textbf{selecting the top 5\% of tokens with the highest entropy.} From this candidate set, we filter for tokens that explicitly signal a logical transition or a new phase of reasoning (e.g., "thus," "however," "first"). These markers indicate the boundaries of distinct reasoning steps. The reason we can do this is because it can be argued that for large language models, the first tokens within each steps has the highest entropy, making it a reliable source for classifying the steps. The math arguement for this statement  is at  \ref{A} in appendix \ref{appendix1}
.

Segmentation is performed at these markers, subject to two constraints to prevent excessive granularity: \textbf{maintaining a minimum interval between segments and aligning segmentation points with complete sentence boundaries.} This ensures that each resulting step is a coherent and meaningful unit of reasoning.

\subsubsection{Lightweight Approximate Attribution Metric}
% From an information-theoretic perspective, every correct reasoning step should provide relevant information that aids in predicting outcome A. If a reasoning step no longer increases the information about A, it indicates that the step is ineffective. Specifically, correct steps should enhance the predictive information about Y (with positive information gain), while incorrect steps yield no gain or even reduce information. When the model backtracks to the correct path, the information gain of subsequent steps turns positive, reflecting progress in effective reasoning.Based on this theory, we designed a step-wise attribution reward. By characterizing the causal measurement of different segmented steps on the final answer, we further focus on optimizing our advantage gradients. This reward can indirectly guide a diversified attribution optimization strategy, which refines gradient objectives through a dynamic advantage function while balancing exploration and exploitation.In essence, this step-wise information gain is equivalent to the conditional mutual information: $I(S_j; Y | {S}_{<j})$.

From an information-theoretic perspective, an effective reasoning step should increase the predictive information about the final answer, $Y$. Steps that provide no new information are considered ineffective.

Based on this principle, we designed a step-wise attribution reward to measure the causal impact of each reasoning step on the final outcome. This reward refines our advantage gradients, guiding the model to prioritize paths with positive information gain. Essentially, this step-wise gain is the conditional mutual information $I(S_j; Y | {S}_{<j})$.

We measure the contribution of a reasoning step $S_j$ to the answer Y using conditional mutual information $I\left(S_j ; Y \mid S_{<j}\right)$, where a high value indicates a critical step and a low value suggests redundancy.

\textbf{Mutual Information Measurement via Entropy Approximation}:
To compute this efficiently, we leverage the relationship between mutual information and conditional entropy:
\begin{equation}
I\left(S_j ; Y \mid S_{<j}\right)=H\left(S_j \mid S_{<j}\right)-H\left(S_j \mid S_{<j}, Y\right)
\end{equation}
This equation establishes that the conditional entropy of a step, \(H(S_j | S_{<j})\) is an upper bound on its information gain. Consequently, if a step's conditional entropy is near zero, its information gain must also be negligible, making the step redundant. This agrees with the entropy approach that we use in classifying the steps. 

In practice, we use a judge model(in our case, the model under training itself) to approximatge the mutual information gain, by computing the following quantity:

\begin{align}
    \mathcal{C}_{\text{attr}}(S_i)=L
(S_1,...S_i,Y)-L(S_1,S_2,...,S_{i-1},Y)
\end{align}
It can be intuitively understood as the amount of information $S_i$ brings to the final answer $Y$, which in our case, is the .

\subsubsection{Attribution Advantage}
To assign credit from a sparse outcome reward to individual reasoning steps, we introduce an \textbf{Attribution Exploration Reward}. This reward shaping technique modulates an entropy bonus for each step to dynamically balance exploration and exploitation.

Instead of applying a uniform entropy regularizer, our approach is differentiated based on the step's contribution (its advantage) and the model's confidence (its entropy).

\paragraph{Positive Advantage (Helpful Steps)}
When a step contributes positively to the correct answer, we encourage useful exploration.
\begin{itemize}
    \item \textbf{For high-entropy (low-confidence) steps:} We increase the reward bonus to explore diverse yet effective reasoning paths.
    \item \textbf{For low-entropy (high-confidence) steps:} We keep the bonus minimal to exploit the known-good strategy.
\end{itemize}

\paragraph{Negative Advantage (Harmful Steps)}
If a step is detrimental to the outcome, we suppress unhelpful exploration. In this case, we reduce the entropy bonus to encourage the model to generate more coherent and predictable sequences, preventing further deviation.

Our objective is to modulate reward signals based on a step's contribution (attribution) and the model's uncertainty (entropy). This allows for targeted exploration, preventing the entropy collapse seen with uniform reward schemes. The advantage function for a step $S_i$ is defined as:
\begin{equation}
\mathcal{A}_{\text{attr-diversity}}(S_i) = \mathcal{A}_{\text{base}} \cdot \mathcal{C}_{\text{attr}}(S_i) \cdot w(H(S_i | S_{<i}), \mathcal{C}_{\text{attr}}(S_i), \mathcal{A}_{\text{base}})
\end{equation}
where:
\begin{itemize}
    \item $\mathcal{A}_{\text{base}}$ is the base advantage signal, indicating if the trajectory leads to a correct ($>0$) or incorrect ($<0$) final answer.
    \item $\mathcal{C}_{\text{attr}}(S_i)$ is the attribution score, quantifying the importance of step $S_i$ to the outcome.
    \item $H(S_i | S_{<i})$ is the conditional entropy of the step, measuring model uncertainty.
    \item $w(\cdot)$ is a dynamic weight function that implements our differentiated entropy regulation.
\end{itemize}

The core of our method lies in the weight function $w(\cdot)$, which adjusts the reward based on three distinct scenarios:
\begin{equation}
w(\cdot) = \begin{cases} 
1 + \beta \cdot \frac{H - H_{\min}}{H_{\max} - H_{\min}} & , \text{if } \mathcal{C}(S_i) \geq \theta \text{ and } \mathcal{A}_{\text{base}} > 0 \\ 
1 - \gamma \cdot \frac{H - H_{\min}}{H_{\max} - H_{\min}} & , \text{if } \mathcal{C}(S_i) < \theta \text{ and } \mathcal{A}_{\text{base}} > 0 \\ 
1 - \gamma \cdot \frac{H - H_{\min}}{H_{\max} - H_{\min}} & , \text{if } \mathcal{A}_{\text{base}} < 0 
\end{cases}
\end{equation}
In summary, the logic is:
\begin{enumerate}
    \item \textbf{For helpful and important steps}, we increase the reward for high-entropy outputs (controlled by $\beta$) to encourage exploration of diverse, effective reasoning paths.
    \item \textbf{For helpful but unimportant steps}, we penalize high-entropy outputs (controlled by $\gamma$) to maintain coherence.
    \item \textbf{For harmful steps}, we penalize high-entropy outputs (controlled by $\gamma$) to suppress unproductive exploration.
\end{enumerate}

Finally, this term is integrated into a base RL advantage function via a hyperparameter $\alpha$:
\begin{equation}
\mathcal{A}^{\text{final}} = \mathcal{A}^{\text{base}} + \alpha \mathcal{A}^{\text{attr-diversity}}
\end{equation}
This method provides fine-grained, step-level reward modulation, in contrast to prior work (e.g., GRPO, DAPO) that applies a uniform advantage value across all tokens in a response.

\subsubsection{Progressive Curriculum Learning Strategy}
\begin{figure}[htbp]
    \centering
    \includegraphics[width=1.0\textwidth, keepaspectratio]{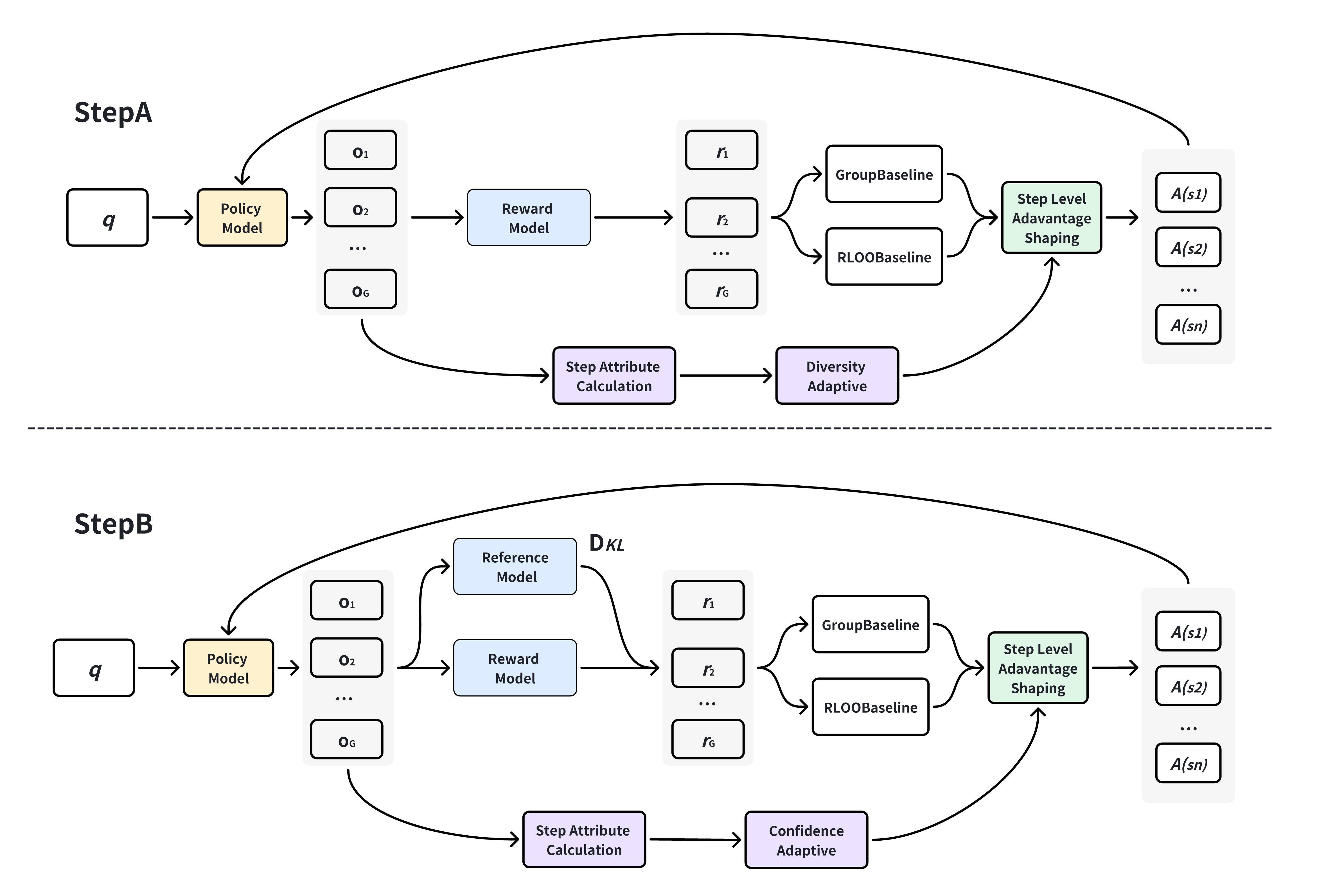}
    \caption{The Two-Stage ACPO Training Framework}
    \label{fig:acpo_stages}
\end{figure}

To effectively manage the exploration-exploitation trade-off in RL, we introduce a two-stage curriculum learning strategy for our ACPO algorithm. Each stage has a distinct objective, moving from broad exploration to focused convergence.

\subsection*{Stage 1: Broad Exploration}
The initial stage aims to maximize exploration to discover diverse and effective reasoning paths. To achieve this, we employ a KL-free objective that removes constraints on the policy, allowing it to explore a wider solution space.

The reward mechanism is based on step-level attribution. Within each reasoning step, the reward is distributed with a \textbf{uniform weight} to every token. For more difficult problems, we use hierarchical sampling with a higher temperature to further boost output diversity. The objective is:
\begin{align*}
    \mathcal{J}_{A C P O}(\theta)=&\mathbb{E}_{\left[q \sim P(Q),\left\{o_j\right\}_{j=1}^G \sim \pi_{\text {old }}(O \mid q)\right]} \\
&\left.\left.\frac{1}{G} \sum_{j=1}^G \frac{1}{\left|o_j\right|} \sum_{t=1}^{\left|o_j\right|}\left\{\min \left[\frac{\pi_\theta\left(o_{j, t} \mid q, o_{j,<t}\right)}{\pi_{\text {old }}\left(o_{j, t} \mid q, o_{j,<t}\right)} \mathcal{A}_{j, t}^{\text {contribute }}\right), \operatorname{clip}\left(\frac{\pi_\theta\left(o_{j, t} \mid q, o_{j,<t}\right)}{\pi_{\text {old }}\left(o_{j, t} \mid q, o_{j,<t}\right)}, 1-\varepsilon, 1+\varepsilon\right) \mathcal{A}_{j, t}^{\text {contribute }}\right)\right]\right\} \\
&
\mathcal{A}_{j, t}^{\text {contribute }}=\left(\hat{\mathcal{A}}_{j, t}+\alpha \mathcal{A}_{j, t}^{\text {attr }- \text { div }}\right)
\end{align*}

where:
\begin{itemize}
    \item \textbf{$A_{i, t}$}: The estimated advantage for generating token $t$ in output trajectory $i$.
    
    % \item \textbf{$P(Q)$}: The probability distribution over the set of all problems $Q$.
    
    \item \textbf{$\left\{o_i\right\}_{i=1}^G$}: A set of $G$ output trajectories sampled from the old policy $\pi_{\text{old}}$.
    
    \item \textbf{$\pi_\theta(o_{i, t} \mid q, o_{i,<t})$}: The probability of generating token $o_{i, t}$ given the problem $q$ and preceding tokens $o_{i,<t}$, according to the current policy $\pi_\theta$.
    
    \item \textbf{$\pi_{\text{old}}(o_{i, t} \mid q, o_{i,<t})$}: The probability of generating token $o_{i, t}$ under the old policy $\pi_{\text{old}}$.
    
    \item \textbf{clip function}: A function that constrains the policy update ratio to the range $[1-\epsilon, 1+\epsilon]$ to prevent destructively large updates.
    
    \item \textbf{$\epsilon$}: The clipping threshold, a hyperparameter defining the update constraint range.
    
    \item \textbf{$\beta$}: The weighting coefficient for the KL divergence penalty.
    
    % \item \textbf{$D_{\mathrm{KL}}(\pi_\theta \| \pi_{\mathrm{ref}})$}: The Kullback-Leibler (KL) divergence, measuring the difference between the current policy $\pi_\theta$ and a reference policy $\pi_{\text{ref}}$.
\end{itemize}

\subsection*{Stage 2: Targeted Convergence}
In the second stage, we shift from exploration to exploitation, aiming to refine the strategies discovered in Stage 1. The key changes are:
\begin{enumerate}
    \item \textbf{KL-Divergence Constraint:} We introduce a KL-divergence penalty to stabilize training and ensure the policy does not deviate too far from the effective policies found in the first stage ($\pi_{\text{ref}} = \pi_{\text{Stage 1}}$).
    \item \textbf{Confidence-Weighted Rewards:} The reward allocation is refined. Instead of uniform weighting, we re-weight rewards within each step to prioritize \textbf{high-confidence tokens} (those with higher generation probabilities). This encourages the model to commit to its most certain and correct reasoning steps.
\end{enumerate}
This combined approach accelerates convergence while ensuring self-consistency. The objective function is updated to:

\begin{equation}
\begin{split}
    \mathcal{J}_{A C P O}(\theta) = &\mathbb{E}_{\substack{
                                    q \sim P(Q), \\
                                    \{o_j\}_{j=1}^G \sim \pi_{\text{old}}(O \mid q)
                                  }}
    \Bigg[  \frac{1}{G} \sum_{j=1}^G \frac{1}{|o_j|} \sum_{t=1}^{|o_j|} \bigg\{ \\
    & \min \bigg( \frac{\pi_\theta(o_{j, t} \mid q, o_{j,<t})}{\pi_{\text{old}}(o_{j, t} \mid q, o_{j,<t})} \mathcal{A}_{j, t}^{\text{contribute}},  \operatorname{clip}\left(\frac{\pi_\theta(o_{j, t} \mid q, o_{j,<t})}{\pi_{\text{old}}(o_{j, t} \mid q, o_{j,<t})}, 1-\varepsilon, 1+\varepsilon\right) \mathcal{A}_{j, t}^{\text{contribute}} \bigg)- \mathbb{D}_{\mathrm{KL}}(\pi_\theta || \pi_{\mathrm{ref}}) \bigg\} \Bigg]
\end{split}
\end{equation}

\begin{align}
    \mathcal{A}_{j, t}^{\text {contribute }} &= \left(\hat{\mathcal{A}}_{j, t}+\alpha \mathcal{A}_{j, t}^{\text {attr-div }}\right) \cdot \left(1+\exp\left(\log \pi_\theta\left(o_{j, t} \mid q, o_{j,<t}\right)\right)\right) \\
    \mathbb{D}_{\mathrm{KL}}\left(\pi_\theta || \pi_{\mathrm{ref}}\right) &= \frac{\pi_{\text {ref }}\left(o_{j, t} \mid q, o_{j,<t}\right)}{\pi_\theta\left(o_{j, t} \mid q, o_{j,<t}\right)}-\log \frac{\pi_{\text {ref }}\left(o_{j, t} \mid q, o_{j,<t}\right)}{\pi_\theta\left(o_{j, t} \mid q, o_{j,<t}\right)}-1
\end{align}
where   \textbf{$D_{\mathrm{KL}}(\pi_\theta \| \pi_{\mathrm{ref}})$} is the K3 estimator of 
Kullback-Leibler (KL) divergence, measuring the difference between the current policy $\pi_\theta$ and a reference policy $\pi_{\text{ref}}$.
% \begin{description}
%     \item[$P(Q)$] The probability distribution over the set of all problems $Q$.
    
%     \item[$q \sim P(Q)$] A single problem $q$ sampled from the distribution $P(Q)$.
    
%     \item[$\pi_{\text{old}}$] The old policy, used to generate sample data.
    
%     \item[$\pi_{\theta}$] The current policy being optimized, parameterized by $\theta$.
    
%     \item[$\left\{o_i\right\}_{i=1}^G$] A set of $G$ output trajectories sampled from the old policy, i.e., $o_i \sim \pi_{\text{old}}(\cdot \mid q)$.
    
%     \item[$o_{i, t}$] The $t$-th token (action) in the $i$-th output trajectory $o_i$.
    
%     \item[$\pi_\theta(o_{i, t} \mid q, o_{i,<t})$] The probability of generating token $o_{i, t}$ given the problem $q$ and the preceding tokens $o_{i,<t}$, according to the current policy $\pi_\theta$.
    
%     \item[$A_{i, t}$] The estimated advantage for generating token $o_{i, t}$ in its corresponding context.
% \end{description}

\section{Experiments}

We present the evaluation to validate the effectiveness of our proposed ACPO algorithm. In this initial phase, we use the Qwen2.5-Math-7B model \cite{qwen_team_2025_qwen25} as our foundation and conduct experiments within the Open-r1 framework to demonstrate the core benefits of our approach.

\subsection{Experimental Setup}

\subsubsection{Datasets and Benchmarks}
In terms of data, we integrated guided data and raw data. First, we leverage the LLM that we would like to train to perform reasoning on the dataset, and define the reasoning difficulty based on the number of successful rollouts. Subsequently, for data with higher difficulty, we conduct multiple rollouts, select reasoning trajectories that yield correct outcomes, and extract questions and prefixes from these trajectories to construct new sample pairs. This approach enables dataset fusion guided by an expert knowledge base.

For RL training, we utilize the DAPO-17k dataset, which comprises $
17,000$ high-quality mathematical problems, with trajectories generated $2,000$ problems To evaluate model performance, we use four standard mathematical reasoning benchmarks: AIME 2024 (MAA, 2024), AIME 2025 (MAA, 2025), AMC 2023 (MAA, 2023), and MATH500\cite{hendrycks2021measuringmathematicalproblemsolving}.

\subsubsection{Evaluation Metrics}
We report performance using the Acc@8 metric, which measures the percentage of problems solved correctly within the top 8 generated responses. During evaluation, all responses are generated using a sampling temperature of 0.8 and a top-p value of 0.95.

\subsubsection{Implementation Details}
All experiments are conducted using the Qwen2.5-Math-7B model (Yang et al., 2024). Our RL implementation is built on the TRL library, with GRPO  serving as the primary baseline. The baseline is configured with a KL divergence loss coefficient, $\beta$, of 0.

Our training hyperparameters include a learning rate of $1\times10^{-6}$ 
  and a global batch size of 192. During the trajectory rollout phase, we generate eight responses for each prompt with a sampling temperature of $1.0$ and a top-p of $0.95$. All models are trained on 8 NVIDIA H20 GPUs, leveraging gradient checkpointing and BF16 precision for computational efficiency.

\subsection{Main Results}

The primary results of our experiments are presented in the figures below. We provide a comparative analysis on the AIME 2024 benchmark, focusing on key metrics that highlight the impact of our ACPO framework. Specifically, the figures illustrate the policy entropy, average response length, and Acc@8 score, comparing our method against the baseline. 

\begin{figure}[h]
  \centering   
  \begin{subfigure}[b]{0.3\textwidth} % [b]表示底部对齐
    \centering
    \includegraphics[width=\linewidth]{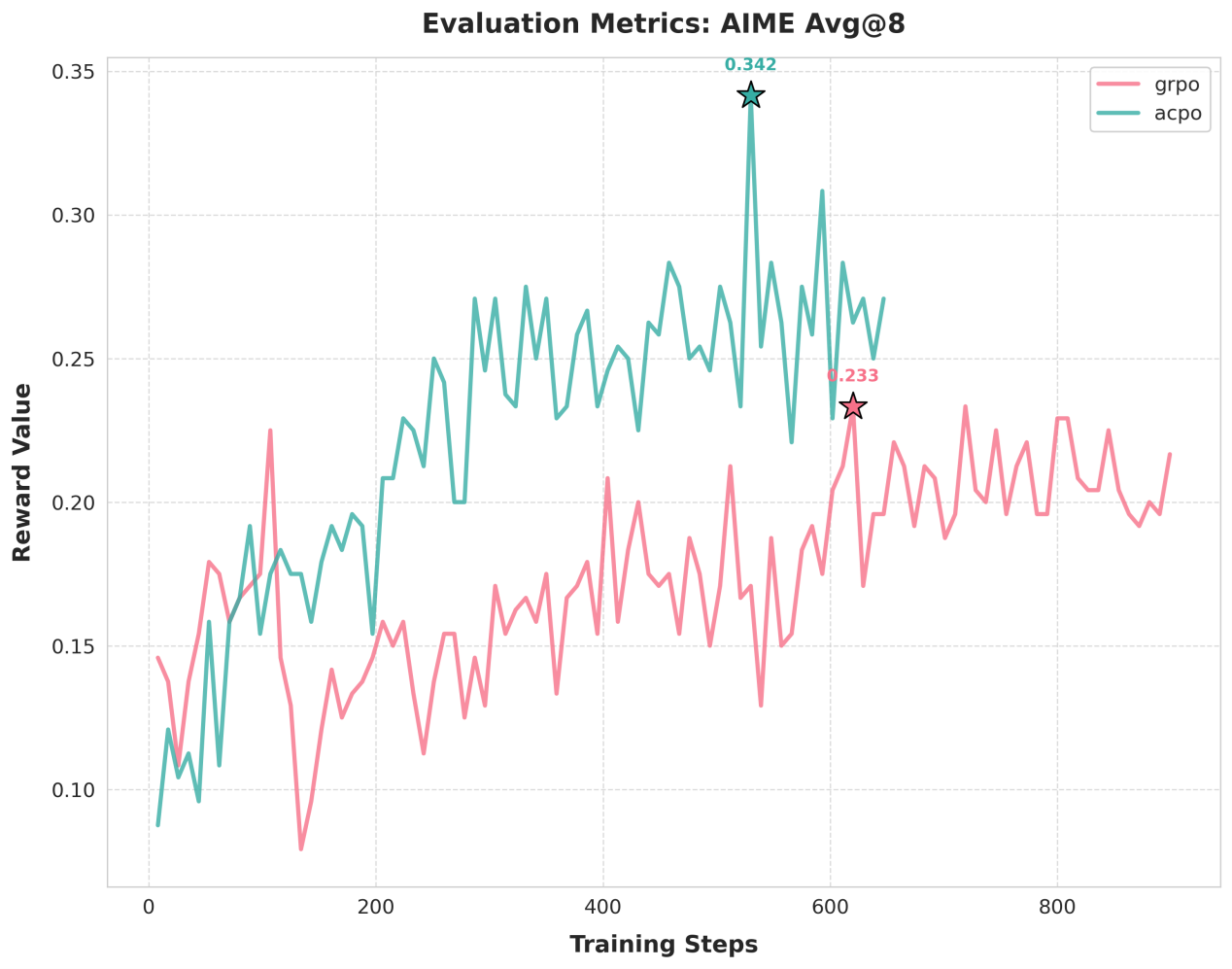} 
    \caption{Reward curve}
    \label{fig:sub1}
  \end{subfigure}
  \hfill 
  \begin{subfigure}[b]{0.3\textwidth} 
    \centering
    \includegraphics[width=\linewidth]{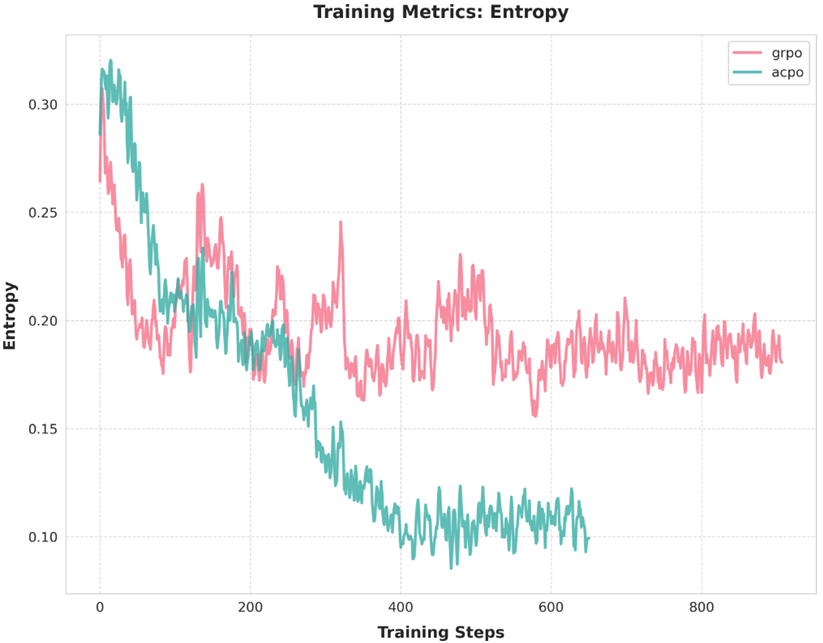}
    \caption{Entropy}
    \label{fig:sub2}
  \end{subfigure}
  \begin{subfigure}[b]{0.3\textwidth} 
    \centering
    \includegraphics[width=\linewidth]{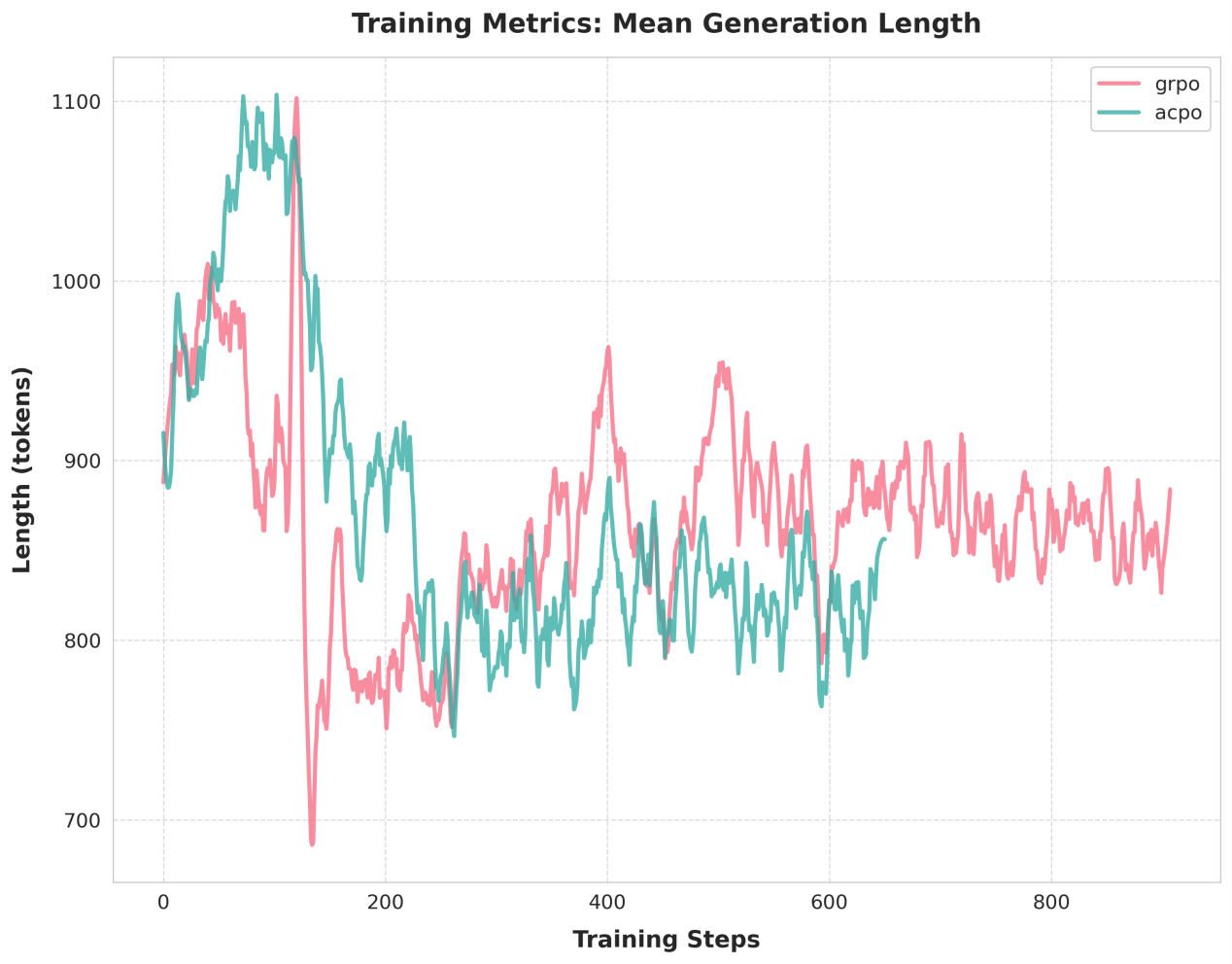}
    \caption{Average output tokens}
    \label{fig:sub2}
  \end{subfigure}
  \caption{train results}
  \label{fig:main}
\end{figure}

\begin{table}[htbp]
\centering
\begin{tabular}{lccccc}
\toprule
\textbf{Evaluation set} & \textbf{AIME2024} & \textbf{AIME2025} & \textbf{AMC23} & \textbf{Math-500} & \textbf{Average} \\
\midrule
Qwen2.5-Math-7B-base & 15.2 & 7.5 & 42.77 & 57.6 & 30.7675 \\
GRPO & 23.3 & 12.9 & 64.5 & 78.6 & 44.825 \\
ACPO & 34.2 & 16.25 & 71.9 & 83.4 & 51.4375 \\
\bottomrule
\end{tabular}
\caption{Performance comparison on various math evaluation sets.}
\label{tab:math_performance}
\end{table}
As illustrated in the figures, the ACPO algorithm effectively enhances the capabilities of the base model compared to the original GRPO algorithm, boosting the score on the AIME2024 dataset from $23.3$ to $34.1$. The entropy patterns also reveal that during the early stages of training, ACPO maintains a high entropy level due to its guided exploration strategy. Subsequently, in the later stages, the entropy decreases to a lower value as ACPO identifies critical steps and applies targeted optimization. The output length is also moderately reduced compared to the original GRPO algorithm.

The table shows a comparison between our method and current state-of-the-art approaches. On the Qwen2.5-Math-7B model, our method achieves an average improvement of $20\%$, which demonstrates its effectiveness.

\section{Conclusion}
In this paper, we introduced ACPO, a novel step-level reinforcement learning framework designed for fine-grained credit assignment in language models. ACPO's primary contributions are threefold: 1) a method for \textbf{effective step classification using entropy}; 2) a mechanism for \textbf{step-grained advantage attribution} to enable precise reward allocation; and 3) an \textbf{exploration-centric two-stage training curriculum}. Our experiments demonstrate that ACPO achieves state-of-the-art performance on mathematical reasoning benchmarks, validating that a more precise, verifiable credit assignment mechanism leads to more robust and effective policies.

Unlike methods that rely on structured prompting to delineate reasoning steps—an approach that can be rigid and may constrain model performance—ACPO learns to identify these steps organically from the generation process. This intrinsic approach provides a more flexible and generalizable solution for complex reasoning tasks.

Several promising avenues for future research remain. One direction is to explore even finer-grained credit assignment and step-refinement techniques. For instance,  we merge the classified steps based on \textbf{token surprisal}. The surprisal of a step's initial tokens indicates its logical importance: High surprisal signals a critical logical transition or new information, reflecting the model's uncertainty. These steps are preserved as distinct units. Low surprisal suggests a redundant continuation or grammatical filler, generated with high certainty. its full effectiveness as a standalone mechanism warrants deeper investigation.  Furthermore, we used alternative methods for estimating the mutual information of a step could be explored, such as using the \textbf{perplexity} of the answer; a higher perplexity in the remaining answer after removing a step might indicate a higher information contribution from that step. 

Additionally, future work could focus on developing reward mechanisms at the token level by identifying the functional role of each token (e.g., ``knowledge token," ``reasoning token" and ``exploration token"). A deeper understanding and strategic use of negative samples will also be crucial for enhancing RL training efficiency. Finally, while this work focused on mathematical reasoning, applying and evaluating ACPO across a broader range of domains is an important next step to confirm its generalizability.

\bibliography{iclr2025_conference}
\bibliographystyle{iclr2025_conference}

%%%%%%%%%%%%%%%%%%%%%%%%%%%%%%%%%%%%%%%%%%%%%%%%%%%%%%%%%%%%%%%%%%%%%%%%%%%%%%%
%%%%%%%%%%%%%%%%%%%%%%%%%%%%%%%%%%%%%%%%%%%%%%%%%%%%%%%%%%%%%%%%%%%%%%%%%%%%%%%
% APPENDIX
%%%%%%%%%%%%%%%%%%%%%%%%%%%%%%%%%%%%%%%%%%%%%%%%%%%%%%%%%%%%%%%%%%%%%%%%%%%%%%%
%%%%%%%%%%%%%%%%%%%%%%%%%%%%%%%%%%%%%%%%%%%%%%%%%%%%%%%%%%%%%%%%%%%%%%%%%%%%%%%
\newpage
\appendix
\onecolumn
\section{Math proofs } \label{appendix1}

ACPO is fundamentally grounded in the notion of discrete reasoning "steps." Recent work by \cite{ton2025understanding} establishes that such steps are not only theoretically meaningful but can also be empirically captured and analyzed. Given this theoretical foundation, the practical challenge lies in developing robust methods to classify these steps.

First, we introduce some useful concepts in information theory:

\begin{definition}[Mutual information]
    The mutual information of one part of $A_i$ and the other section $A_j$ are defined as difference of the entropy $H(A_i)$ and the conditional entropy $H(A_i|A_j)$.
    \begin{align}
        I(A_i,A_j)&=H(A_i)-H(A_i | A_j)\nonumber \\
        &=H(A_i)+H(A_j)-H(A_i,A_j) \nonumber \\
        &=H(A_j)-H(A_j|A_i),
    \end{align}
    \begin{align}
        & H(A_i)=-\sum_j p_i(x_j)\log (p_i(x_j)),\\
         &H(A_i,A_j)=-\sum_{a,b}p_{i,j}(x_a,x_b)\log p_{i,j}(x_a,x_b),\\ &H(A_i|A_j)=-\sum_{a,b}p(x_a,x_b)\log(x_a|x_b).
    \end{align}
    $p_i$ is the probability distribution of the $i-th$ step. and $p_{i,j}$ is the union probability distribution of the $i$-th and $j$-th step.
\end{definition} 
It's understood with instinct: The mutual information of $A_j$ and $A_j$ are the amount of uncertainty reduced of one with the prior knowledge of the other, and should be symmetric. And we have:

\begin{theorem}\label{A}
    $\min(H(A_i),H(A_j)) \geq I(A_i,A_j) \geq 0. $
\end{theorem}

Also, for a step of $m$ tokens  $T_1, T_2,...T_m$, suppose that:
\begin{align*}\label{assumption}
    &H(T_1)\geq H(T_2), H(T_2|T_1)\geq H(T_3|T_2),\\
    &H(T_3|T_1,T_2)\geq H(T_4|T_2,T_3),\\
    &...,\\
    &H(T_{n-1}|T_{1},T_{2},...T_{n-2})\geq H(T_{n}|T_{2},T_{3},...T_{n-1})
\end{align*}

then we have :

\begin{theorem}\label{B}
\begin{align}
    &H(T_1)\geq H(T_2|T_1) \geq H(T_3|T_1,T_2)\geq...\nonumber\\ 
    &\geq H(T_3|T_1,T_2)
\end{align}
\end{theorem}

Remarks: The assumption is important and also natural in autoregressive models like LLMs because it reveals autoregressive nature of the distributions,since:
\begin{align*}
    P(T_2=t)=\sum_p P(T_2=t|T_1=p)P(T_1=p).
\end{align*}
The conditional distribution $P(T_2=t|T_1=p)$ is statistically sharper than the original distribution. Thus making $H(T_1)\geq H(T_2)$. Other assumptions work in  similar ways.

Now we give proofs of \ref{A} and \ref{B}:

\begin{proof}[proof of \ref{A}]
    We just notice that from definition that $p_i(x_a)= \sum_b p_{i,j}(x_a,x_b)$:
    \begin{align*}
        H(A_i)&=-\sum_a p_i(x_a) \log p_i(x_a)\\
        &=\sum_a \sum_b p_{i,j}(x_a,x_b)\log(p_i(x_a))
    \end{align*}
    Thus:
    \begin{align*}
        I(A_i,A_j)&=H(A_i)+H_(A_j)-H(A_i,A_j)\\
                &=\sum_{a,b}p(x_a,y_b)\log \frac{p(x_a,y_b)}{p_i(x_a)p_j(y_b)}.
    \end{align*}
which can be seen as a generalized KL divergence of two variables, so they are non-negative. On the other hand, we can prove it using Jensen's inequality that since $\log (x)$ is a concave function, we have:
\begin{align*}
&-I(A_i,A_j)=-\sum_{a,b}p(x_a,y_b)\log \frac{p(x_a,y_b)}{p_i(x_a)p_j(y_b)}\\&=\sum_{a,b}p(x_a,y_b)\log \frac{p_i(x_a)p_j(y_b)}{p(x_a,y_b)}\\
&\leq \log \left(\sum_{a,b}p(x_a,y_b)*\frac{p_i(x_a)p_j(y_b)}{p(x_a,y_b)}\right) \\
&=\log 1=0
\end{align*}
Thus $I(A_i,A_j)\geq 0$. On the other hand, we know from definition that $H(A_i|A_j)\geq0$, then we know that $I(A_i,A_j)=H(A_i)-H(A_i|A_j) \leq H(A_i)$. Same holds for $A_j$. Combining everything together, we finished our proof.
\end{proof}

As for \ref{B}. This follows  from the premise  that  $H(T_1)\geq H(T_2)\geq H(T_2|T_1)$, and $H(T_2|T_1)\geq H(T_3|T_2) \geq H(T_3|T_2,T_1)$. Other inequalities follows in the same way.

The reason for introducing both of these is because they play a sigfinicant role in practically detecting the contributions of the steps to the final answer within our entropy approach. 

Suppose we have an unclassified answer consisting of tokens $T_1,T_2,...,T_n$ To properly track and evaluate the importance of its constituent steps, we can use entropy as a tool to identify them. Based on our observations from \ref{B}, the first tokens of each step often exhibit the highest entropy due to an information advantage.

This entropy-based classification provides a guideline for determining the number of steps. Once we have successfully classified all the steps, we need to evaluate their respective contributions. This is crucial because traditional methods often assume all steps have equal weight, when in fact, some steps are significantly more important than others. We can evaluate the specific value of each step effectively.

% For instance, if we have an answer to a math problem with prompt $P$, steps $S_1,S_2,...S_n$ and a final answer $A$. We can measure the importance of the steps, by considering the mutual information for these steps, for example $I(S_t,A|P,S_1,S_2,..., S_{t-1})$. And the more information it has, the bigger contribution it makes for the final answer, the more rewards it get assigned to during the training process

% You can have as much text here as you want. The main body must be at most $8$ pages long.
% For the final version, one more page can be added.
% If you want, you can use an appendix like this one.  

% The $\mathtt{\backslash onecolumn}$ command above can be kept in place if you prefer a one-column appendix, or can be removed if you prefer a two-column appendix.  Apart from this possible change, the style (font size, spacing, margins, page numbering, etc.) should be kept the same as the main body.
%%%%%%%%%%%%%%%%%%%%%%%%%%%%%%%%%%%%%%%%%%%%%%%%%%%%%%%%%%%%%%%%%%%%%%%%%%%%%%%
%%%%%%%%%%%%%%%%%%%%%%%%%%%%%%%%%%%%%%%%%%%%%%%%%%%%%%%%%%%%%%%%%%%%%%%%%%%%%%%

\end{document}